\documentclass{amia}
\usepackage{graphicx}
\usepackage[labelfont=bf]{caption}
\usepackage[superscript,nomove]{cite}
\usepackage{color,soul}
\usepackage{siunitx}
\usepackage[table]{xcolor}
\usepackage{amsfonts}
\usepackage{bm}
\usepackage{subcaption}
\usepackage{enumitem}

\begin{document}

\title{Leveraging Clinical Time-Series Data for Prediction: A Cautionary Tale}
\author{Eli Sherman$^{1}$, Hitinder Gurm, MD$^{2}$, Ulysses Balis, MD$^{3}$, Scott Owens, MD$^{3}$, Jenna Wiens$^{1}$}

\institutes{
    $^1$University of Michigan Computer Science and Engineering, Ann Arbor, MI; $^2$Michigan Medicine Cardiology, Ann Arbor, MI; $^3$Michigan Medicine Pathology, Ann Arbor, MI;\\
}

\maketitle

\noindent{\bf Abstract}
\textit{In healthcare, patient risk stratification models are often learned using time-series data extracted from electronic health records. When extracting data for a clinical prediction task, several formulations exist, depending on how one chooses the time of prediction and the prediction horizon. In this paper, we show how the formulation can greatly impact both model performance and clinical utility. Leveraging a publicly available ICU dataset, we consider two clinical prediction tasks: in-hospital mortality, and hypokalemia. Through these case studies, we demonstrate the necessity of evaluating models using an outcome-independent reference point, since choosing the time of prediction relative to the event can result in unrealistic performance. Further, an outcome-independent scheme outperforms an outcome-dependent scheme on both tasks (In-Hospital Mortality AUROC .882 vs. .831; Serum Potassium: AUROC .829 vs. .740) when evaluated on test sets that mimic real-world use.}

\section{Introduction}
The widespread adoption of electronic health records (EHR) for collecting and analyzing patient data presents a promising avenue for improving patient care\cite{hersh2004health}. These retrospective data have been used in traditional statistical analyses to identify relationships between patient data and patient outcomes. For example, extensive work has been done to identify risk factors for sepsis\cite{pierrakos2010sepsis}, hospital readmission\cite{kansagara2011risk}, and heart failure-induced mortality\cite{alba2013risk}, among other conditions. In such analyses, the primary goal is to increase our understanding of the relationship between covariates and the outcome of interest. In contrast, the increasing accessibility of machine learning (ML) approaches has triggered a shift in focus to the development (and deployment) of predictive models, where the primary goal is good predictive performance \cite{chia2012looking,paxton2013developing}. 

This shift in focus necessitates a similar shift in methodology. For example, traditional analyses are often limited to a small number of variables, based on expert-driven clinical hypotheses or observations. Because the risk of over-fitting is low, researchers in this setting rarely consider a held-out test set. In a high-dimensional setting a held-out test set is imperative. More clinical researchers are recognizing this need and adapting their analysis appropriately, but this alone is not enough. The precise problem statement also requires an important, but often overlooked, reformulation. When the goal is prediction, the choice of method by which one chooses to extract one's data is critical, for both training and testing purposes.

In particular, one must be careful to evaluate candidate models in a way that accurately estimates how they will perform when applied in clinical practice. This stands in contrast to previous analyses, in which the models may not have been intended for predictions. In a predictive setting, \textit{when} a model will be applied, i.e., when a prediction will be made, is important. Two common methods for defining the time of prediction include: 1) indexing relative to a clinical event of interest e.g., onset of infection, and 2) indexing relative to an outcome-independent fiducial marker e.g., time of admission. The first approach is most common in a traditional retrospective analysis. In such settings, examples are typically derived by considering the event of interest and looking backwards to extract data collected prior to the event\cite{ray2016shock,debiane2014utility}. The second approach must be used a predictive analysis setting. Here, examples are derived based on a fiducial marker that precedes the clinical event of interest but is temporally independent of that event e.g., a particular operation/procedure.

In this paper, through two case studies, we illustrate some of the subtleties that surround prediction tasks and demonstrate how using an evaluation method that does not reflect clinical practice can result in misleading results. Leveraging a large publicly available dataset of ICU admissions, we apply several indexing techniques to two prediction tasks. In one prediction task, we aim to predict in-hospital mortality. In the other prediction task, we aim to predict hypokalemia, low serum potassium. We consider these two tasks since they are both important from a clinical perspective, but also because they present a contrast in terms of the number of events per admission. I.e., for the first task we may observe several test results, but in the second task there is only a single endpoint - in-hospital mortality. Through these experiments we illustrate the importance of carefully defining one's problem setup and demonstrate how evaluation performance can vary across settings. 

In the following sections, we will: 1) discuss previous work in the development of predictive tools including examples of several indexing techniques, 2) formalize different indexing techniques/problem setups and outline how we will apply them to our case studies, 3) present results on the two real-world prediction tasks, and finally 4) reflect on the implications our experiments have on choosing an indexing method for extracting training and evaluation sets for a prediction model.

\section{Related Work}
Traditional low-dimensional statistical analyses are common in the clinical literature. Such studies typically focus on modeling relationships between covariates and outcomes\cite{ray2016shock,debiane2014utility, schadendorf2015pooled, loi2014tumor, grana2014longitudinal}. While the focus of these statistical analyses is often on testing hypotheses about relationships among variables, such approaches have been applied to prediction tasks. The popular mortality prediction scoring systems APACHE III\cite{knaus1991apache} and SAPS II\cite{moreno2005saps} serve as illustrative examples of this approach. In both models, a relatively small number of candidate covariates are hand-selected by experts and a mapping between covariates and outcomes is learned using standard statistical methods (e.g., logistic regression). This approach is limited in that data are collected at a specific point in the admission (24 hours after the time of admission, and at the time of ICU transfer for APACHE and SAPS respectively) and thus the models are only designed to make a single prediction.

In recent years, the focus has shifted to developing prediction models with high-dimensional feature spaces using machine learning techniques. These models are applied throughout the admission, providing updated predictions. As such, the question of how to index and extract time-series data from the EHR is now critically important. Numerous examples of indexing from the time of admission (or a related fiducial point) exist in both the healthcare and machine learning literature, including applications to patient status estimation\cite{georgatzis2016input}, diagnosis\cite{choi2015doctor}, and sepsis\cite{henry2015targeted}.

Despite the number of examples in which data are indexed based on an outcome-independent fiducial marker, the approach of indexing based on the event of interest still creeps into analyses today (perhaps since it was so common in a traditional, non-prediction centric, setting). Oftentimes, identifying the use of the event-based method is subtle and it is necessary to carefully consider the features included in the proposed model to confirm its use. For example, in work on learning a model for \textit{Clostridium difficile} infection (CDI)  risk prediction, it is not explicitly stated how examples are extracted\cite{dubberke2011development}. However, in previous CDI work event-based indexing is clearly used\cite{dubberke2007clostridium}. Another paper developed a risk stratification tool for predicting sepsis risk in ICU patients in which the model used length of stay as a feature \cite{back2016development}. These types of features serve as a clear indication of backward-looking example selection. In another example, a paper proposing a heart failure prediction system generated covariates by looking backwards, testing different prediction intervals, from the diagnosis \cite{wang2015early}. Collectively, these examples present an issue of data leakage, allowing extra information into the training process through problematic indexing schemes.

Others have published work on the potential pitfalls of working with EHR data\cite{paxton2013developing}, and developed tutorial-style overviews of the process of developing and validating a clinical prediction tool\cite{labarere2014derive}. These papers provide a fairly comprehensive discussion of how to undertake careful data analysis to create useful prediction tools. In contrast to previous work, we focus entirely on how training and test data are indexed when developing clinical prediction models.

\section{Methods}\label{sec:Methods}
In this section, we present and contrast common approaches for extracting and indexing EHR data. We begin by describing the general framework for data extraction, highlighting key choices one must make when extracting data. Then, we present the two case studies in which we aim to predict laboratory test results (specifically hypokalemia, i.e., low serum potassium) and in-hospital mortality.

\subsection{Problem Setup and Notation}
We limit our discussion to clinical prediction tasks during a hospital admission. For an admission, we have a set of $d$ irregularly sampled, timestamped features. Additionally, we have a set of irregularly sampled, timestamped outcomes $(y_j, t_j)_{j=1}^k$ where $y_j$ is recorded at time point $t_j$. Here, we restrict $y \in \{0,1\}$. Note that it may be possible for an outcome to occur multiple times, only a single time (e.g., death) or never at all within a given admission (i.e., $k=0$). In the following sections, we will describe procedures for mapping these features and outcomes for an admission $i$ into feature vector-outcome pairs of the form $(\mathbf{x}, y)$ where feature vector $\mathbf{x} \in \mathbb{R}^{d}$ represents the covariates used to predict the outcome variable $y$. Since our focus is prediction, $\mathbf{x}$ represents data recorded prior to observing outcome $y$.

\subsection{Indexing Longitudinal Data for Prediction}
\begin{figure}[t]
\centering
\includegraphics[scale=.5]{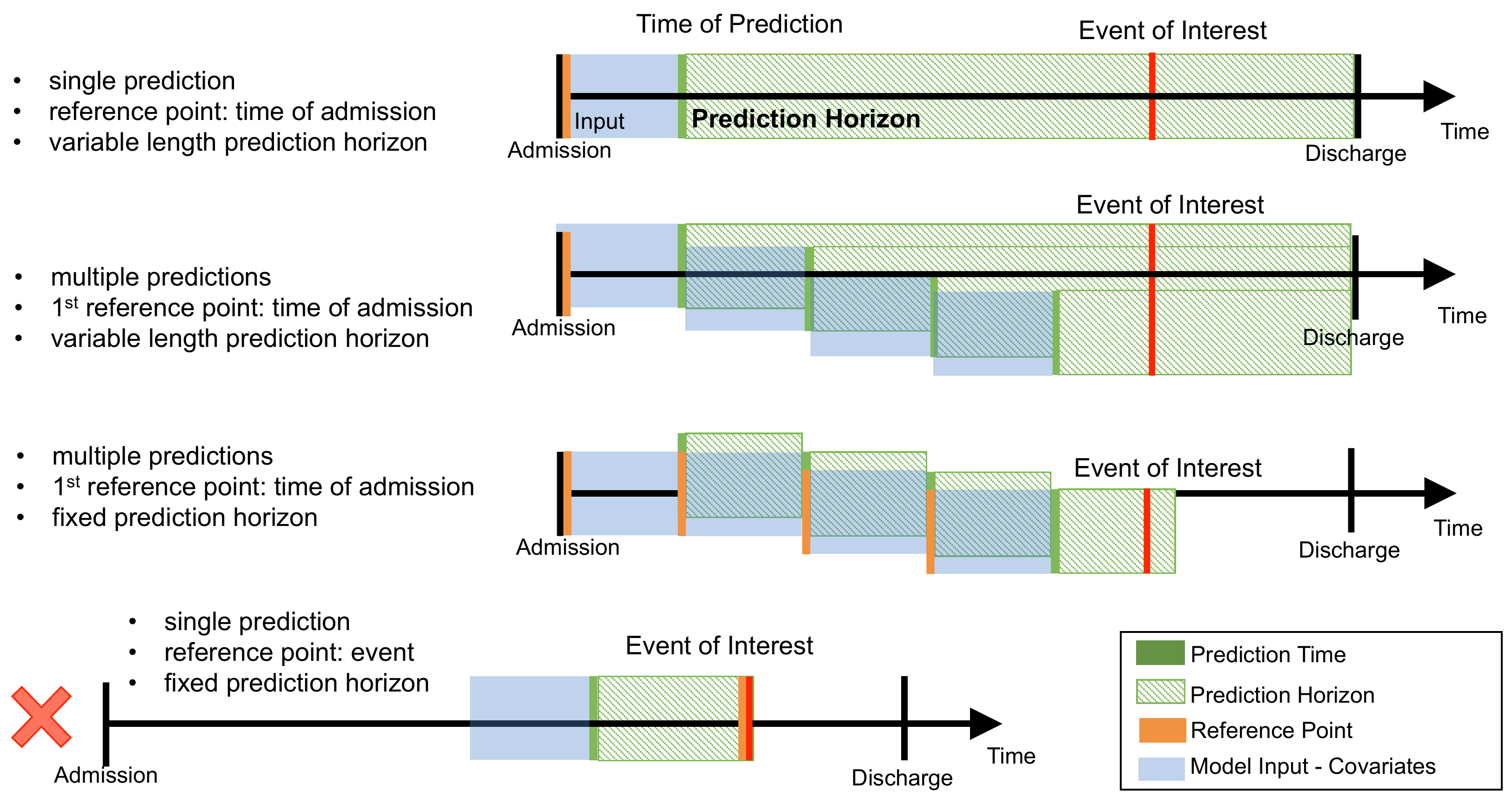}
\caption{Sample time series' depicting the various ways that data sets can be extracted. The first three correspond to varying numbers of predictions per outcome measurement. The final one uses the event of interest as the reference point which we note should \textit{not} be used for model evaluation since it does not mirror any clinical use case.}
\label{fig:indexing}
\end{figure}

Considering the raw data for each admission as a collection of clinical data, we now outline careful considerations for extracting $(\mathbf{x}, y)$ pairs. Figure \ref{fig:indexing} serves as a guide for the following descriptions.

\begin{itemize}[leftmargin=*]
\item\textbf{Time of Prediction} The time of prediction, $t_p$ is the time point corresponding to when a predictive model is applied to the data. The time of prediction affects covariate extraction and outcome/label extraction (more on this below), and is determined relative to some reference point $t_0$. Given a model that predicts a particular outcome, one may choose to apply it at a single time point (as in the first time series in Figure \ref{fig:indexing}), resulting in a single prediction per admission. If such a setup is desired then one test example should be extracted for each admission based on an outcome-independent fiducial marker (e.g., time of admission). Alternatively, one may choose to apply the model multiple times throughout the admission, updating predictions as new data become available (as in the second and third time series' in Figure \ref{fig:indexing}). To mimic this setup, examples can be extracted on a rolling basis, again starting from an outcome-independent reference point $t_0$. More formally, starting from $t_0$, one may extract either a single example $\mathbf{x}_p$ representing the admission from the time of admission to the time of prediction $t_p$, or multiple examples $(\mathbf{x}_1,t_1)$, $(\mathbf{x}_2,t_2)$, ... $(\mathbf{x}_m,t_m)$ (at perhaps regularly spaced intervals). In the latter, each feature vector is updated based on data available at the time of prediction. The corresponding outcome variable $y_p$ depends on the prediction horizon and the availability of ground truth, discussed next. 

\item\textbf{Prediction Horizon} 
The prediction horizon $h$ (or window) is defined as the period for which a prediction applies and aids in defining $y_p$. It begins at the time of prediction $t_p$. In some settings, the prediction horizon is fixed and remains constant across examples, (e.g., when predicting 30-day mortality, or the value of a laboratory test in the next 12 hours; see the third time series in Figure \ref{fig:indexing}). In many cases, the event of interest occurs at most once during the prediction horizon (e.g., 24-hr mortality), but is observed continuously (i.e., at every time point we know whether or not the patient is alive). In such settings ground truth is readily available and the corresponding $y_p$ can simply be set to the observed value. In other settings, however, ground truth is only available at specific time points (e.g., when a laboratory test was ordered and returned). Here, it may be necessary to impute the corresponding outcome value or label. In the case study described below, we use a ``copy-and-hold'' approach: if there is no ground truth at the target time, we use the most recently observed $y$ value that precedes the target time. Another way around this issue is to simply adjust the prediction horizon to be near zero and the prediction time to be driven the laboratory test order. In many settings, the time of the laboratory test order is still an outcome independent reference point, but depending on the task this may have limited clinical utility. Finally, in some settings, the prediction horizon may vary across admissions. E.g., many researchers aim to predict in-hospital mortality. Here, the prediction horizon varies with the length of stay, but $y_p$ can be extracted similarly as if it were fixed.

\end{itemize}

Once the time of prediction and prediction horizon are defined, one can extract a feature vector $\mathbf{x}_p$ based on data available at the time of prediction, and $y_p$ based on observations made during the prediction horizon. This results in either a single $(\mathbf{x}, y)$ pair per admission in the test set, or $m_i$ pairs for each admission $i$.

To ensure evaluation accurately reflects how a model will perform in a practice, test data must be extracted in a way that mimics the clinical use case. In particular, when extracting test data it is necessary to use an outcome-independent reference point since outcomes are not available at test time. However, when training, we may want to make use of this additional information. In particular, one may choose to define the time of prediction based on the time of the event (e.g., work backwards from time of death) when extracting examples. When making multiple predictions for each admission, one extracts test examples regularly throughout each admission. But, applied to the training set, this procedure could introduce bias, since patients with longer stays will appear more often. To mitigate this issue, one can re-sample the data. E.g., for each admission in the training set, one can randomly select (with replacement) $k$ examples from the full set of examples. Through a series of experiments presented in the next section, we illustrate the effects these choices can have on the predictive performance of the learned model. 

\subsection{Dataset \& Prediction Tasks}

To measure the effects of the different data extraction approaches on the predictive performance of a model, we considered two prediction tasks applied to the same ICU dataset. In particular, we leverage the MIMIC-III Database\cite{johnson2016mimic}. This dataset consists of $58,976$ ICU admissions collected at a Beth Israel-Deaconess Medical Center in Boston, MA over the course of 12 years. The median  length of stay is 6.9 days. We consider variables related to patient demographics, medications, laboratory tests, vital signs, and fluid inputs and outputs that are present in at least 5\% of admissions. We encode each patient's demographics as binary features. For the other variables, we generate summary statistics (e.g., min. value, max. value, mean value based on the several hours preceding prediction time) to encode the features for a given example. Using these data, we consider two clinical prediction tasks:

\textbf{Predicting In-hospital Mortality} In-hospital mortality prediction is a well studied problem in the healthcare-related literature. It is frequently used to benchmark new methods. For this task, we include all adult hospital admissions, resulting in $49,909$ admissions. Here, we aim to predict whether or not the patient will expire over the course of the remainder of the admission (i.e., a variable prediction horizon).

\textbf{Predicting Hypokalemia} Accurate predictions of whether or not a particular laboratory test will return a hypokalemic result could help clinicians make better informed decisions regarding whether or not to order that test. This could in turn lead to a reduction in the number of unnecessary laboratory test orders. Here, we aim to predict hypokalemia. We focus on serum potassium (as opposed to other laboratory tests), since it is one of the most high volume tests in U.S. healthcare\cite{weydert2005simple, miyakis2006factors}. From the MIMIC database, we include all adult patients with at least one serum potassium test. This results in a final study population of $49,354$ admissions.

Using these data, we aim to predict whether or not the serum potassium laboratory test will return a hypokalemic result, defined as serum potassium $< 3.5$, within the next 12 hours (i.e., our prediction horizon is 12 hrs.). This threshold was derived from a standard reference range. The prediction horizon was chosen based on the fact that many laboratory tests are ordered on a standard 24-hour schedule and discretionary testing occurs outside of this cycle. A 12-hour prediction window provides information that would ordinarily correspond to a discretionary test and therefore has the potential for eliminating the need for an extra discretionary test. Later, we consider another setup in which we generate a prediction of the current hypokalemia risk every time a potassium test is ordered. We describe this setup in more detail in Section 4.3.

\subsection{Experimental Setup}
Across all experiments, we build our training and test sets by repeatedly (100 times) randomly assigning 25\% of admissions to the test set, and the remaining 75\% to the training set. We are careful to not allow any admission to be represented in both the training set and test set since this would allow our learning algorithm to ``memorize" the admission in question and subsequently perform better on the examples it had already seen. We use 5-fold cross validation on the training set to select hyper-parameters and retrain on the entire training set using the optimized parameters. 

We extract examples according to the particulars of each experiment. Each example is a feature vector with $1,222$ features that encode summary information about the patient's status over the 12 hours preceding prediction time. For instance, the patient's glucose measurements over the 12 hours preceding the prediction time are encoded into variables describing the minimum, mean, and maximum values over the 12 hours. This summarization step allows us to account for the fact that different patients might have a different number of tests over the course of those 12 hours. We pick 12 hours so as to focus on the most recent clinical data while capturing temporal trends.

Using the training data, we train a classification model. Since the focus is on the problem setup, rather than the overall classification performance, we use a simple linear approach, specifically the liblinear implementation of Linear SVMs\cite{fan2008liblinear}. For each experiment, we report mean and standard deviation statistics for area under the receiver operating curve (AUROC) scores across all 100 train-test splits. For models in which we make multiple predictions, each with a variable horizon as in the in-hospital mortality prediction task, we include each patient only once in the AUROC calculation. For each patient, we have a series of predictions, each in the form of a real valued number, the output of our classifier's decision function. We sweep the decision threshold and if any of a patient's predictions exceed that threshold we classify that patient as positive, and negative otherwise. This procedure is necessary since we are making predictions with a variable prediction horizon.

\section{Experiments \& Results}
In this section, we describe and present results for several experiments using various problem setups. We analyze how different reference points and different numbers of predictions per admission impact model performance.

\subsection{Event-dependent Training and Testing Yields Misleadingly Good Performance}
We first illustrate the importance of selecting an appropriate outcome-independent reference point when extracting test data. For each task, we consider two different reference points: 1) time of admission and 2) the event of interest. 

We first consider a single prediction setup for each task. For the in-hospital task, we use two different reference points to extract data. We 1) use the time of admission as our reference point, making a prediction 12 hours after admission, and 2) use the time of death (or time of discharge in the case of negative examples) as our reference point, making a prediction 24 hours prior to the end of the admission. When extracting examples for the laboratory testing task we 1) use time of admission as our reference, making predictions every 12 hours and 2) use the time of each potassium test as our reference, with a time of prediction 12 hours prior to each serum potassium test. 

For these and all experiments in this paper, when we use the time of admission as our reference point, our first prediction is made 12 hours after the time of admission.

In order to isolate the impact that using an admission-based reference point has on this problem (rather than the number of predictions), we also consider a multiple prediction formulation. We create multiple-prediction training sets for the in-hospital mortality task using both event-independent and event-dependent indexing. For the event-independent training set we make predictions regularly -- every 24 hours -- throughout the admission. For the event-dependent training set, we make predictions in 24 hour increments, looking backwards from the end of the admission all the way to the beginning of the admission. The AUROC is calculate as described previously, where each patient contributes equally.

\begin{table}[h!]
\centering
    \begin{tabular}{@{}|l | l | r | r|>{\bfseries}S[table-format=3.2]@{}}
 \hline
 Task & Reference Point & N Examples &N Pos. Examples\\
 \hline
\texttt{In-Hosp.} \texttt{Mort.}&Admission&49,909&5,244\\
\texttt{In-Hosp.} \texttt{Mort.}&Event&49,909&5,244\\
\texttt{Hypokalemia}&Admission&2,831,268&256,891\\
\texttt{Hypokalemia}&Event&644,371&69,234\\
\hline
\end{tabular}
\caption{Number of extracted examples and the number of positive examples for the single mortality and hypokalemia prediction tasks. Note that the two in-hospital mortality sets have the same number of examples as they represent making a single prediction per admission, 12 hours after admission and 24 hours before death/discharge respectively.}
\label{tab:single_set_sizes}
\end{table}

\begin{table}[h!]
\centering
    \begin{tabular}{@{}|l | l | r | r|>{\bfseries}S[table-format=3.2]@{}}
 \hline
 Task & Reference Point & N Examples &N Pos. Examples\\
 \hline
\texttt{In-Hosp.} \texttt{Mort.}&Admission&504,147&58,984\\
\texttt{In-Hosp.} \texttt{Mort.}&Event&476,170&54,338\\

\hline
\end{tabular}
\caption{Number of examples extracted and the number of positive examples for each multiple prediction mortality set.}
\label{tab:multi_set_sizes}
\end{table}

The resulting number of examples for each task and reference point are given in Tables \ref{tab:single_set_sizes} (single prediction tasks) and \ref{tab:multi_set_sizes} (multiple prediction tasks). Admission-based reference points typically result in more examples compared to event-based reference points. The effects of these differences are explored later in this section.

\begin{figure}[t]
\centering
\includegraphics[scale=.4]{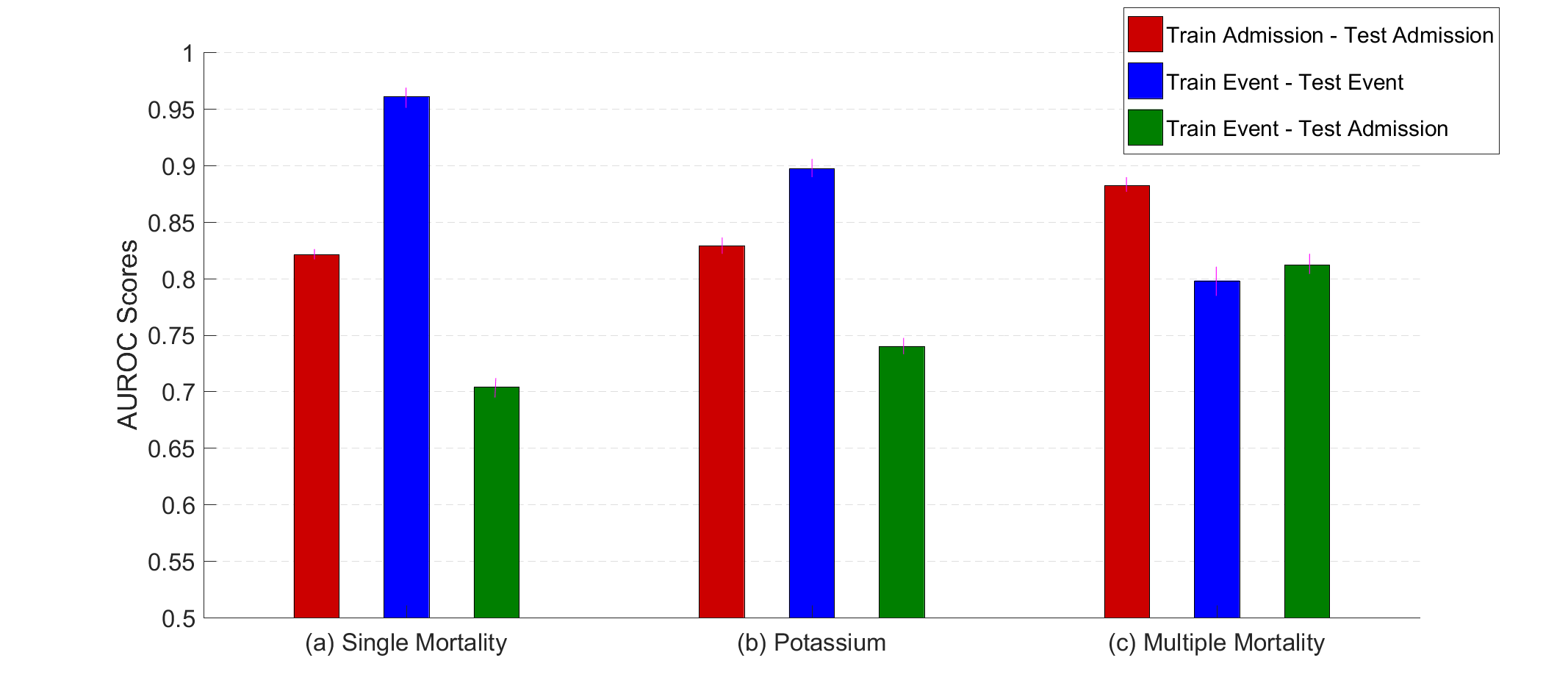}
\caption{AUROC scores for the single mortality prediction, hypokalemia prediction, and multiple mortality prediction setups. The legend indicates the training set and test set schemes for each bar. Errors are denoted with purple lines at the top of each bar. Observe that for the single mortality and hypokalemia tasks, the experiment using both an event-dependent training set and an event-dependent test set yields very high performance indicating the clear bias present in those formulations.}
\label{fig:auroc}
\end{figure}

The results for these experiments are presented in Figure \ref{fig:auroc}. The three sections of this figure give results for the single mortality (a), potassium (b), and multiple mortality (c) tasks respectively. The red bars correspond to admission-based training sets applied to admission-based test sets; the blue bars correspond to event-based training sets applied to event-based test sets; and the green bars correspond to event-based training sets applied to admission-based test sets.

The first thing to note is the extremely good performance apparent in (a) and (b) when both training and test data are extracted in an outcome-dependent manner (AUROC .963 and .897 for in-hospital mortality and serum potassium respectively). However, when such a model is applied to outcome-independent test data (green), performance is significantly worse. Using time of event as a reference point for evaluation is not reflective of a realistic clinical use case. This conclusion is critically important and motivates our use of outcome-independent evaluation sets for the experiments that follow.

For the multiple prediction model (c), the event-dependent model performs relatively poorly when applied to both the event-dependent test set and to the event-independent training set. We suspect that this model does not perform as well as the single prediction model (applied to the event-dependent test sets) because the single prediction model is tested on only the easiest cases, those immediately preceding the end of the admission. The multiple prediction model, on the other hand, has a wider variety of examples (for both training and testing). While the point-wise performance of this model on the event-dependent test set is lower than the performance on the event-independent test, this difference is not statistically significant.
 
These results demonstrate the importance of admission-based indexing. When testing, admission-based indexing more accurately reflects the desired clinical use case. Moreover, in this more realistic test setting, a model trained using admission-based indexing outperforms a model developed using event-based indexing.

\subsection{Training Data Must Be Chosen Carefully to Avoid Biased Models}
When we used time of admission as the reference point for in-hospital mortality prediction above, we made daily predictions of in-hospital mortality. This approach is potentially problematic since there are patients who are represented in the training set several times more than average due to longer hospital stays. Here, we consider how sub-sampling the training data can affect performance.

We explore this issue by comparing the performance of the multiple predictions admission-based training set to a sub-sampled training set derived from the admission-based set. We randomly sample 6 times (with replacement) from each patient's full set of training examples. This number was chosen because it corresponds to the median number of days in an admission and we are building a daily prediction model. We round down from 6.9 because our first prediction is not generated until after $12$ hours have passed in the admission, as described above. This ensures that each patient is equally represented in the training set.

\begin{table}[h!]
\centering
    \begin{tabular}{@{}|l | r|>{\bfseries}S[table-format=3.2]@{}}
 \hline
Sampling Scheme&Mean (STD) AUROC\\
 \hline
Median Sample&.900 (.003)\\
All Samples&.882 (.005)\\
 \hline
\end{tabular}
\caption{AUROC scores for the admission-based sub-sampling experiment. All Samples corresponds to a multiple prediction admission-based model. Median sample is a sub-sampling (based on the median number of days in an admission) of All Samples. Both are evaluated on a test set that uses time of admission as the reference point.}
\label{tab:sampling_aurocs}
\end{table}

The results for this experiment are presented in Table \ref{tab:sampling_aurocs}. From these results we can see that sub-sampling helps performance, improving AUROC from .882 using the original training set, to .900 using the re-sampled set. Sub-sampling reduces the impact of outliers (i.e., patients with extended hospital admissions) on the model, and thus, we are able to learn a more generally applicable model.

\subsection{The Use Case Can Significantly Affect Performance}
As discussed above, evaluation should mimic the intended clinical use case. Different clinical use cases can result in different models and performance can vary greatly. To highlight this point we consider two additional experiments described below.

Many of the most widely used ICU risk scoring systems, including APACHE and SAPS, are defined based on a single time of prediction. While useful for making care decisions in the early stages of the admission, it is not clear how these scores generalize to multiple predictions (updated based on newly available data). To explore this issue, we compare a model for predicting in-hospital mortality that makes only a single prediction at 12 hours to one that makes a prediction every 12 hours. While the model that makes multiple predictions has the advantage of additional data, it also has more opportunity to get the prediction incorrect. To compare these two approaches, we test both models identically in a multiple prediction setting (variable prediction horizon). 

\begin{table}[h!]
\centering
    \begin{tabular}{@{}|l | l | r|>{\bfseries}S[table-format=3.2]@{}}
    \hline
Model Type&Evaluation&Mean (STD) AUROC\\
 \hline
Single Prediction&Rolling&.821 (.009)\\
Multiple Prediction&Rolling&.882 (.005)\\
 \hline
\end{tabular}
\caption{Single and multiple mortality prediction AUROC scores. Single prediction is trained using data available 12 hours into the admission. Multiple Prediction is the previously described multiple prediction admission-based mortality model. Both models are tested on an admission-based test set with predictions made on a rolling basis throughout the admission.}
\label{tab:sing_mult_aurocs}
\end{table}

We observe that a training set that makes predictions throughout the admission outperforms a training set that only makes a single prediction at the beginning of the admission (Table \ref{tab:sing_mult_aurocs}). A model trained on a single prediction made near the time of admission cannot necessarily be applied to the remainder of the admission. The model benefits from additional examples collected throughout the admission.

For the problem of serum potassium testing, we considered a use case in which one aims to predict future laboratory values every 12 hours. One could imagine a different use case in which one aims to predict the ``current'' laboratory test value. In this setting, a prediction is made every time a serum potassium test is ordered and the prediction horizon is immediate or essentially zero. To measure performance in this setting, we extract the training and test sets identically: we use each observed serum potassium result in a patient's admission as the time of prediction and make a prediction using the covariates available prior to this order. The observed result serves as ground truth. In total we extract $649,949$ examples, $69,723$ of which are positive.

This extraction scheme is not the same as using an outcome-dependent reference point since the clinical use case is one of ``on-demand'' estimation rather than advance prediction. Note that this model is not directly comparable to an outcome-independent approach. Nevertheless, we can still draw some conclusions about the performance of such an on-demand model (Table \ref{tab:on_demand}).

\begin{table}[h!]
\centering
    \begin{tabular}{@{}|l | r|>{\bfseries}S[table-format=3.2]@{}}
    \hline
Model&Mean (STD) AUROC\\
 \hline
On-Demand &.738 (.003)\\
Time of Admission&.829 (.002)\\
 \hline
\end{tabular}
\caption{AUROC scores for the on-demand and time of admission reference point models on the serum potassium task. Note that these problems are very different and their performance isn't necessarily directly comparable.}
\label{tab:on_demand}
\end{table}

From the results, we note that the on-demand task is more difficult, despite the shorter prediction horizon. This is somewhat expected since we only make predictions when a laboratory test is ordered, versus every 12 hours. The fact that a test was ordered suggests the clinician suspects the value to differ from that of previous values. Again, while the tasks are similar (``predict potassium values'') the resulting performance is not comparable since test data are extracted in different ways. 

While, the on-demand setting does not require imputation (since ground truth is always available), it could fail to mimic a real clinical use case. Once made available, clinicians may query the model more often (e.g., every 12 hours). This would shift the underlying distributions, resulting in a difference in performance.

\section{Summary \& Conclusion }
In this work, we described methods for extracting examples from longitudinal clinical data for prediction tasks. These methods vary in terms of \textit{when} predictions are made (e.g., in reference to what, and how often), and for \textit{how long} each prediction applies (e.g., for the remainder of the admission or for the next 24 hours). These decisions should be largely dictated by the clinical use case. We illustrate how these decisions affect extracted data and in turn model performance using two different prediction tasks: hypokalemia prediction and in-hospital mortality prediction.

In both cases, we showed that using an outcome-dependent point of reference yields misleadingly good performance. Creating a test set from such a point of reference amounts to picking the easier examples and does not accurately reflect how the model would perform in practice. For example, a patient about to be discharged from the hospital will almost certainly be characterized by a healthier set of covariates just prior to discharge than a patient who is about to expire. For evaluation purposes, it is imperative that test examples be extracted independently from the label/outcome. Still, while predictions generated by models trained on outcome-dependent examples are biased, they can help uncover relationships between covariates and outcomes and shed light on our understanding of disease processes.

This work serves as a cautionary tale of the importance of carefully extracting examples from clinical data for the purposes of building a predictive model or clinical decision support tool. When training and testing prediction models using retrospective data, careful attention to the problem setup such that it accurately reflects the intended real-world use is critical. Neglecting to do so could result in a dangerous misinterpretation of the model's clinical value.

\section*{Acknowledgements}
This research program is supported by the Frankel Cardiovascular Center and the Department of Pathology at the University of Michigan, and the National Science Foundation (NSF award numbers IIS-1553146). The views and conclusions in this document are those of the authors and should not be interpreted as necessarily representing the official policies, either expressed or implied, of NSF.

\bibliographystyle{vancouver}
\bibliography{indexing_clinical_timeseries_amia}

\end{document}